\documentclass[11pt]{article}

\usepackage[preprint]{acl}
\usepackage{times}
\usepackage{latexsym}
\usepackage{inconsolata}

\usepackage{booktabs}
\usepackage{multirow}
\usepackage{makecell}
\usepackage{float}
\usepackage{placeins}

\usepackage{listings}

\usepackage[most]{tcolorbox}

\newtcolorbox{promptbox}[1][]{
  enhanced,
  breakable,
  colback=gray!10!white,
  colframe=gray!50!black,
  title=#1,
  colbacktitle=black!70,
  coltitle=white,
}

\usepackage{microtype}      
\usepackage{graphicx}       
\usepackage{booktabs}       
\usepackage{multirow}
\usepackage{amsmath,amssymb}
\usepackage{enumitem}       
\usepackage[nameinlink,noabbrev]{cleveref} 
\usepackage{siunitx}        

\usepackage{tikz}
\usepackage{pgfplots}
\pgfplotsset{compat=1.18}
\usepgfplotslibrary{groupplots}
\usepackage{svg}

\usepackage[T1]{fontenc}
\usepackage{listings}
\crefname{section}{§}{§§}
\Crefname{section}{Section}{Sections}
\usepackage{xspace}

\newcommand{\itermodel}{\textsc{HDSR}\xspace}
\newcommand{\itermodelfull}{\textsc{Hallucination Detection guided Self-Refinement}\xspace}
\newcommand{\model}{\textsc{HDSR-PL}\xspace}

\newcommand{\SynFac}{\textsc{SynFac-Edit}\xspace}

\newcommand{\MimicIV}{\textit{MIMIC-IV-Note v2.2}\xspace}

\newcommand{\HalluGenDi}{\textit{Hallucination-Generated-DI}\xspace}

\newcommand{\modelheader}[1]{\multicolumn{6}{l}{\textit{#1}}\\}

\usepackage{colortbl}
\usepackage{xcolor}
\usepackage{pgf}

\def\vmin{0}
\def\vmax{40}

\newcommand{\heatcell}[1]{
  \pgfmathparse{max(min((#1-\vmin)/(\vmax-\vmin),1),0)}
  \edef\colorpercent{\pgfmathresult}
  \cellcolor{blue!\the\numexpr100-\colorpercent*100\relax!orange!60}#1
}

\usepackage{xcolor}
\usepackage{soul}   
\sethlcolor{yellow}

\title{Hallucination Detection-Guided Preference Optimization for Clinical Summarization}

\author{
  \textbf{Shamanth Kuthpadi Seethakantha}$^{1*}$ \quad \textbf{Dung Ngoc Thai}$^{2*}$ \quad \textbf{Vara Prasad Gudi}$^{1*}$ \\
  \textbf{Simran Tiwari}$^2$ \quad \textbf{Rami Matar}$^3$ \quad \textbf{Avijit Mitra}$^1$ \\
  \textbf{Wenlong Zhao}$^1$ \quad \textbf{Andrew McCallum}$^4$ \quad \textbf{Wael Salloum}$^2$ \\[0.5em]
  \small $^{1,4}$Manning College of Information and Computer Sciences \quad $^2$Ensemble HP \quad $^3$Columbia University\\
  \small $^1$\texttt{\{skuthpadi,vgudi,avijit,wenlongzhao\}@umass.edu},  
   $^2$\texttt{\{simran.tiwari,june.thai,wael.salloum\}@ensemblehp.com} \\
  \small $^3$\texttt{\{rhm2142\}@columbia.edu}, $^4$\texttt{\{mccallum\}@cs.umass.edu} \\
  \small $^*$Equal contribution
}

\begin{document}
\maketitle

\begin{abstract}

Large language models (LLMs) have shown promise on summarization tasks, but they often produce hallucinations, which are unsupported or incorrect statements that limit their reliability in specialized healthcare applications. We introduce \itermodelfull (\itermodel), an inference-time method that leverages hallucination detectors to guide iterative summary revisions toward factual corrections. Building on this, we propose \itermodel for Preference Learning (\model), which converts detector-guided refinement trajectories into preference pairs for model finetuning. Extensive experiments show that our methods substantially reduce hallucinations for Llama and Gemma models in summarizing real-world clinical notes from \MimicIV.  For example, \itermodel reduces 24\% and \model reduces 48\% hallucinations in Llama-3.1-8B-Instruct. Importantly, both methods preserve summary fluency, coherence, and relevance according to human expert and LLM-Jury evaluations. Together, these results demonstrate that detection-informed refinement and preference learning offer an automated solution for improving factual faithfulness in clinical summarization.
\end{abstract}

\section{Introduction} 

\begin{figure*}[t]
    \centering
    \includegraphics[
        width=\textwidth,
        clip
    ]{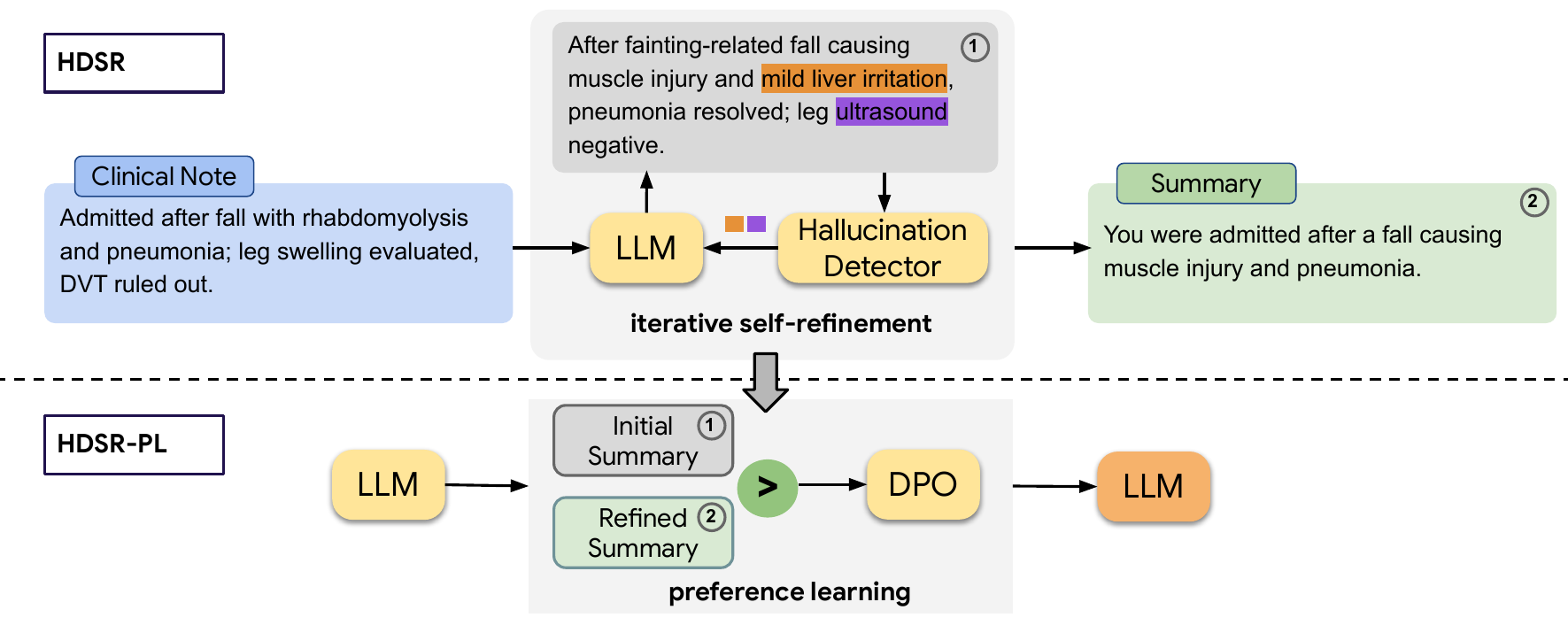}
    \caption{\textbf{Overview of hallucination mitigation via detection-informed self-refinement.} Given an input clinical note, a language model generates an initial summary that may contain unsupported or hallucinated medical content. A hallucination detector identifies unsupported content, which is used to guide iterative self-refinement toward removing factual errors rather than stylistic changes (\textit{top; \itermodel}). The intermediate outputs from \textit{\itermodel} process are converted into preference pairs and used for preference learning (e.g., DPO), amortizing faithful behavior and yielding hallucination-mitigated summaries at inference time (\textit{bottom; \model}). }
    \label{fig:overview}
\end{figure*}

Large language models achieve strong summarization performance but often generate hallucinations, defined as content not supported by the source or inconsistent with world knowledge \citep{maynez2020FaithfulnessFactuality, ji2023SurveyHallucination, tang2023EvaluatingLarge}. Hallucinations remain prevalent even in recent LLMs, with rates of roughly 40-50\% on benchmarks such as \textsc{FaithBench} \citep{bao2025FaithBenchDiverse} indicating persistent factual unreliability. This issue is critical in clinical summarization, where LLMs condense long patient records to support care delivery. Despite comparable performance with clinician‑written summaries \citep{veen2024AdaptedLarge}, hallucinations manifest as fabricated or distorted clinical statements that are hard to detect \citep{asgari2025FrameworkAssess, kim2025MedicalHallucinations}. Even domain‑adapted models frequently introduce hallucinations expressed with medically plausible terminology, requiring expert review \citep{hegselmann2024DataCentricApproach, williams2025EvaluatingLarge, fang2024UnderstandingFaithfulness, das2025HallucinationsKey}, and underscoring the need for stricter factual reliability in clinical summarization than in general‑domain tasks.

Existing hallucination mitigation strategies are broadly divided into training-time and inference-time approaches. Training-time methods include domain pretraining, continual pretraining, supervised fine-tuning, parameter-efficient adapters \citep{veen2024AdaptedLarge,zaretsky2024GenerativeArtificial}, and specialized loss functions or reinforcement learning objectives \citep{fabbri2021SummEvalReevaluating,bao2025FaithBenchDiverse,asgari2025FrameworkAssess}. These methods require large volumes of high-quality domain data \citep{hegselmann2024DataCentricApproach} and depend on factuality metrics that correlate weakly with clinical correctness, limiting their practical impact. Reinforcement learning from human or AI feedback offers greater flexibility, but requires carefully engineered preference data that is difficult to scale in medical settings \citep{lee2024RLAIFVs}. Inference‑time techniques avoid altering model parameters: retrieval‑augmented generation grounds outputs in external documents yet depends on retrieval quality \citep{koopman2023DrChatGPT, wan2024WhatEvidence, kim2025MedicalHallucinations}, while self‑refinement\citep{madaan2023SelfRefineIterative} and verification loops \citep{tang2023EvaluatingLarge} iteratively critique and revise generations, improving factuality at the cost of higher inference overhead and a risk of over‑editing. Recent work combining iterative refinement with alignment shows such approaches can strip salient content or introduce new hallucinations even when fluency is preserved \citep{ji2023MitigatingHallucination}. A closely related approach is \SynFac \citep{mishra2024SYNFACEDITSynthetic}, which generates synthetic edit feedback for preference optimization but relies on predefined error types and external edit models.

Our pipeline combines hallucination detection with iterative self refinement, using detector feedback to guide self refinement and to form preference pairs from the original and revised summaries.

We then train LLMs using direct preference optimization on these preference pairs, producing models that generate more faithful summaries with fewer hallucinations. 
On the MIMIC IV clinical summarization hallucination datasets \citep{hegselmann2024DataCentricApproach}, our method reduces hallucinations by approximately 24\% through self refinement and by up to 48\% after fine tuning, while preserving summary quality and fluency according to both human and LLM-as-judge evaluations, and without incurring additional inference time cost.

\section{Method}

Figure \ref{fig:overview} illustrates our pipeline, which uses hallucination detection feedback to guide summary revision and train models to prefer faithful summaries.

\subsection{Problem Statement}
Given a source document and a generated summary, the goal of hallucination mitigation is to produce summaries that are informative while remaining faithful to the source. We view a source document and its summary as describing clinical facts documented during a patient’s hospital course. A summary is faithful if these facts are supported by the source document. Hallucinations arise when a summary introduces unsupported facts or adds procedures or findings that were not documented, for example when ``leg ultrasound negative'' appears in the summary without support from the source.

\subsection{Hallucination Detection Guided Self-Refinement}

During self-refinement, a hallucination detector is applied to the generated summary to identify content that is unsupported or inconsistent with the source document. Detector feedback highlights unsupported spans or statements, and the model is prompted to revise these parts while preserving supported content (see Appendix \ref{app:self-refinement}). This process focuses revisions on factual corrections rather than stylistic changes.

Detection and revision are applied iteratively. After each revision, the updated summary is re-evaluated by the detector, and the resulting feedback guides subsequent revisions until a fixed iteration limit is reached or no further hallucinations are detected. This detector guided refinement improves factual alignment while maintaining overall fluency.

We use existing hallucination detectors from prior work, specifically MedCat \citep{Kraljevic2021-ln} and prompt based detectors following the MedAlign annotation guidelines \citep{DBLP:conf/aaai/FlemingLHJRTBGS24}. MedCat links clinical concepts in the source document and the summary to biomedical ontologies in order to flag unsupported or missing content. MedAlign defines a clinically motivated taxonomy of hallucination categories, such as unsupported procedures or medications, which we implement as a prompted detector.

\subsection{Preference Learning from Detection Guided Self-Refinement}

Detection-guided self-refinement produces pairs of summaries for each source document, consisting of an initial summary and a revised summary with fewer hallucinations. We treat the revised summary as the preferred output and the initial summary as the non-preferred output, thereby forming preference pairs without human annotation. We train the summarization model using direct preference optimization on these pairs. This objective encourages the model to internalize detector guided factual corrections, amortizing improvements from self refinement into the model parameters and enabling faithful generation at inference time without additional refinement steps.

\section{Experiments} 

\subsection{Experimental Setup}

We study clinical summarization from Brief Hospital Course (BHC) sections to Discharge Instructions (DI) using datasets derived from \MimicIV \citep{hegselmann2024DataCentricApproach}. Experiments are evaluated on the hallucination-annotated subset \HalluGenDi, following the task formulation and annotation guidelines of \citet{hegselmann2024DataCentricApproach}.

We include GPT-5 \citep{OpenAI2025GPT5} zero-shot as a closed-source reference and primarily focus on LLaMA-3.1-8B-Instruct \citep{llama3.1-8b-instruct} as the open-source base model. In addition, we report results for two smaller open-source models, LLaMA-3.2-3B-Instruct \citep{llama3.2-3b-instruct} and Gemma-3-4B-IT \citep{gemma3-4b-it}, evaluated under a limited setting. For LLaMA-3.1-8B-Instruct, we evaluate zero-shot generation, supervised fine-tuning (SFT), detection-informed self-refinement at inference time (\itermodel), and detection-informed preference learning at training time (\model). Detection-informed variants use MedCat \citep{Kraljevic2021-ln} and MedAlign \citep{DBLP:conf/aaai/FlemingLHJRTBGS24} as hallucination detectors.

\paragraph{Evaluation.} We report entity-level hallucination counts and clinician-based human evaluation on summary quality across four main metrics: Consistency, Coherence, Fluency, and Relevance. We follow the protocol of \citet{hegselmann2024DataCentricApproach}, which includes quantitative fine-grained hallucination annotations identifying unsupported and contradicted content, as well as qualitative analysis of summary quality across the same dimensions. Annotations are performed by a team of clinicians blinded to model identity: for the quantitative task, each summary is annotated independently by two annotators (H1 and H2) with adjudication of disagreements, while for the qualitative analysis, two annotators score summaries independently without adjudication.

For LLaMA-3.1-8B-Instruct, human evaluation is conducted in a blind setting with double annotation and adjudication for hallucination labels, forming the basis of the main results in Table~\ref{tab:main_results}. For LLaMA-3.2-3B-Instruct and Gemma-3-4B-IT, due to resource constraints, we report clinician-provided hallucination counts annotated by a single clinician (H1) and replace human qualitative evaluation with automatic LLM-as-judge scores.

\subsection{Main Results}
\label{sec:main_results}

\begin{table*}[t]
\centering
\scriptsize
\setlength{\tabcolsep}{4pt}
\begin{tabular}{l c | c c c c c}
\toprule
\multirow{2}{*}{\textbf{Model / Method}} &
\multirow{2}{*}{\makecell{\textbf{Hallucination}\\\textbf{Count} $\downarrow$}} &
\multicolumn{5}{c}{\textbf{Summary Quality Metrics}} \\
\cmidrule(lr){3-7}
& &
\textbf{Consistency}$\uparrow$ &
\textbf{Coherence}$\uparrow$ &
\textbf{Fluency}$\uparrow$ &
\textbf{Relevance}$\uparrow$ &
\textbf{Average}$\uparrow$ \\
\midrule

\textit{GPT-5} Prompting & 36 & 3.55 & 4.73 & 4.73 & 4.08 & 4.27 \\
\midrule

\modelheader{LLaMA-3.1-8B-Instruct}
Prompting & 29 & 4.08 & 3.83 & 4.05 & 3.23 & 3.79 \\
SFT & 57 & 3.03 & 4.43 & 4.53 & 3.03 & 3.75 \\
\itermodel (best; MedAlign) & 22 & 4.13 & \textbf{4.48} & \textbf{4.53} & \textbf{3.95} & \textbf{4.27} \\
\model (best; MedCat) & \textbf{15} & \textbf{4.40} & 4.28 & 4.05 & 3.90 & 4.16 \\
\midrule

\modelheader{LLaMA-3.2-3B-Instruct}
Prompting & 26 & 4.13 & \textbf{4.58} & \textbf{4.68} & 4.20 & 4.39 \\
\model (best; MedCat) & \textbf{13} & \textbf{4.23} & \textbf{4.58} & \textbf{4.68} & \textbf{4.30} & \textbf{4.44} \\
\midrule

\modelheader{Gemma-3-4b-it}
Prompting & 15 & 4.23 & 4.65 & 4.68 & 4.48 & 4.51 \\
\model (best; MedCat) & \textbf{13} & \textbf{4.33} & \textbf{4.65} & \textbf{4.70} & \textbf{4.63} & \textbf{4.58} \\
\bottomrule
\end{tabular}
\caption{\textbf{Results on \HalluGenDi.} Both variants of our detector-guided approach outperform the zero-shot and SFT baselines. Note that for LLama-3.2-3B-Instruct and Gemma-3-4b-it, we use automatic LLM-as-Judge for summary quality metrics.}
\label{tab:main_results}
\end{table*}


\begin{table*}[t]
\centering
\scriptsize

\begin{minipage}[t]{0.38\textwidth}
\centering
\setlength{\tabcolsep}{2pt}
\renewcommand{\arraystretch}{1.05}

\caption{Distribution of hallucination error types across models.}
\label{tab:analysis}

\resizebox{\linewidth}{!}{%
\begin{tabular}{lrrrrrrrrrrr}
\toprule
\textbf{Model} &
\rotatebox{90}{\textbf{Unsup. Condition}} &
\rotatebox{90}{\textbf{Unsup. Procedure}} &
\rotatebox{90}{\textbf{Unsup. Medication}} &
\rotatebox{90}{\textbf{Unsup. Time}} &
\rotatebox{90}{\textbf{Unsup. Location}} &
\rotatebox{90}{\textbf{Unsup. Number}} &
\rotatebox{90}{\textbf{Unsup. Name}} &
\rotatebox{90}{\textbf{Unsup. Word}} &
\rotatebox{90}{\textbf{Unsup. Other}} &
\rotatebox{90}{\textbf{Contradicted Fact}} &
\rotatebox{90}{\textbf{Incorrect Fact}} \\
\midrule
\multicolumn{12}{l}{\emph{LLaMA-3.1-8B-Instruct}} \\
Prompting & 8 & 2 & 1 & 3 & 4 & 0 & 1 & 2 & 0 & 8 & 0\\
SFT & 15 & 11 & 8 & 3 & 6 & 0 & 3 & 3 & 1 & 7 & 0\\
Self-Refine (No Det.) & 10 & 2 & 6 & 2 & 3 & 3 & 1 & 1 & 1 & 2 & 0\\
\itermodel (w/MedCat) & 7 & 1 & 3 & 4 & 3 & 2 & 2 & 2 & 0 & 1 & 0\\
\itermodel (w/MedAlign) & 4 & 2 & 1 & 3 & 3 & 1 & 1 & 0 & 2 & 5 & 0\\
\model (best; MedCat) & 3 & 1 & 0 & 1 & 2 & 0 & 0 & 0 & 0 & 8 & 0\\
\bottomrule
\end{tabular}%
}
\end{minipage}%
\hfill
\begin{minipage}[t]{0.60\textwidth}
\centering
\setlength{\tabcolsep}{3pt}
\renewcommand{\arraystretch}{1.05}

\caption{\textbf{Impact of detection signal for hallucination mitigation.} Comparison between self-refinement with and without detectors on LLaMA-3.1-8B.}
\label{tab:ablation_results}

\resizebox{\linewidth}{!}{%
\begin{tabular}{l c | c c c c}
\toprule
\multirow{2}{*}{\textbf{Model / Method}} &
\multirow{2}{*}{\makecell{\textbf{Hallucination}\\\textbf{Count} $\downarrow$}} &
\multicolumn{4}{c}{\textbf{Summary Quality Metrics}} \\
\cmidrule(lr){3-6}
& &
\textbf{Consistency}$\uparrow$ &
\textbf{Coherence}$\uparrow$ &
\textbf{Fluency}$\uparrow$ &
\textbf{Relevance}$\uparrow$ \\
\midrule
Self-Refine (No Det.) & 31 & 3.85 & 4.25 & \textbf{4.58} & \textbf{4.20} \\
\itermodel (w/MedCat) & 25 & 4.08 & \textbf{4.55} & \textbf{4.58} & 4.13 \\
\itermodel (w/MedAlign) & \textbf{22} & \textbf{4.13} & 4.48 & 4.53 & 3.95 \\
\bottomrule
\end{tabular}%
}
\end{minipage}

\end{table*}

Table~\ref{tab:main_results} presents the main results on the \HalluGenDi benchmark, comparing zero-shot generation, supervised fine-tuning (SFT), \itermodel, and \model using LLaMA-3.1-8B-Instruct. We also report limited results for LLaMA-3.2-3B-Instruct and Gemma-3-4B-IT. The following analysis focuses on LLaMA-3.1.

\paragraph{SFT amplifies hallucinations.} Supervised fine-tuning on clinician-written references substantially worsens factual alignment and increases hallucinations (57) to nearly twice the zero-shot baseline. Although supervised fine-tuning improves fluency (4.53) and coherence (4.43), consistency drops sharply (3.03). 

\paragraph{\itermodel improves summary quality and mitigates hallucinations.}Our \itermodel yields a substantial improvement over both baselines. Using MedAlign as the detector, \itermodel reduces hallucinations to 22 while simultaneously achieving the best overall human-evaluated summary quality across all dimensions.
\paragraph{\model further mitigates hallucinations.}\model, using MedCat-derived preference pairs, achieves the lowest hallucination count overall (15), corresponding to an approximate 48\% reduction relative to the zero-shot baseline and a 74\% reduction relative to SFT. \model slightly underperforms \itermodel on fluency and coherence. This trade-off reflects the difference between the two variants of our pipeline: \itermodel benefits from direct, entity-specific detector feedback, whereas \model learns a more general preference for factual alignment that may smooth over fine-grained stylistic details but yields factually aligned outputs.

\paragraph{Analyzing the types of hallucinations.}Table ~\ref{tab:analysis} breaks down hallucinations by error type for LLaMA-3.1-8B, revealing that unsupported conditions, procedures, and medications account for the majority of errors in zero-shot and supervised fine-tuned models. SFT amplifies these clinically critical error types, increasing unsupported conditions from 8 to 15 and procedures from 2 to 11. Self-refinement without detection reduces some surface-level errors but leaves substantial unsupported content uncorrected. In contrast, \itermodel variants consistently reduce errors across multiple categories, particularly unsupported conditions and procedures, indicating that the detector guides corrections toward factually grounded revisions.

\paragraph{Impact of detector on hallucination mitigation.} Table ~\ref{tab:ablation_results} shows that incorporating an explicit detection signal reduces hallucination count for \itermodel by 29\% when using MedAlign as the hallucination detector and 19\% when using MedCat.

\section{Conclusion}
We presented two complementary approaches for mitigating hallucinations in clinical summarization using automatic hallucination detection as supervision. \itermodel is an inference-time method that integrates detector feedback into iterative self-refinement, focusing revisions on unsupported or contradicted medical content while preserving grounded information. \model extends this framework to training time by transforming refinement trajectories into preference pairs, enabling direct preference optimization that amortizes factual alignment.

\section*{Limitations}
Our work highlights several areas for further investigation. First, both \itermodel and \model rely on the quality of the hallucination detector used to provide supervision. While we treat detectors as black-box assessors, their errors--such as false positives that signal supported content or false negatives that miss subtle inconsistencies--can limit the effectiveness of refinement and downstream preference learning. Improving detector accuracy or combining multiple complementary detectors may further enhance performance.

In addition, our evaluation focuses on clinical summarization tasks derived from \MimicIV, specifically BHC to Discharge Instructions generation. Although this is a high-impact and clinically relevant setting, the transferability of our approach to other medical note types or non-clinical summarization domains remains to be explored.

Finally, inference-time self-refinement (\itermodel) incurs additional computational cost due to multiple refinement iterations and detector calls, which may limit its applicability in latency-sensitive settings. While \model addresses this by amortizing refinement into training, it requires additional training resources and preference optimization, which may not be feasible in all deployment scenarios.

\bibliography{refs}

@article{asgari2025FrameworkAssess,
  title = {A Framework to Assess Clinical Safety and Hallucination Rates of {{LLMs}} for Medical Text Summarisation},
  author = {Asgari, Elham and {Monta{\~n}a-Brown}, Nina and Dubois, Magda and Khalil, Saleh and Balloch, Jasmine and Yeung, Joshua Au and Pimenta, Dominic},
  year = {2025},
  month = may,
  journal = {npj Digital Medicine},
  volume = {8},
  number = {1},
  pages = {274},
  publisher = {Nature Publishing Group},
  issn = {2398-6352},
  doi = {10.1038/s41746-025-01670-7},
  urldate = {2025-10-07},
  copyright = {2025 The Author(s)},
  langid = {english}
}

@inproceedings{bao2025FaithBenchDiverse,
  title = {{{FaithBench}}: {{A Diverse Hallucination Benchmark}} for {{Summarization}} by {{Modern LLMs}}},
  shorttitle = {{{FaithBench}}},
  booktitle = {Proceedings of the 2025 {{Conference}} of the {{Nations}} of the {{Americas Chapter}} of the {{Association}} for {{Computational Linguistics}}: {{Human Language Technologies}} ({{Volume}} 2: {{Short Papers}})},
  author = {Bao, Forrest Sheng and Li, Miaoran and Qu, Renyi and Luo, Ge and Wan, Erana and Tang, Yujia and Fan, Weisi and Tamber, Manveer Singh and Kazi, Suleman and Sourabh, Vivek and Qi, Mike and Tu, Ruixuan and Xu, Chenyu and Gonzales, Matthew and Mendelevitch, Ofer and Ahmad, Amin},
  editor = {Chiruzzo, Luis and Ritter, Alan and Wang, Lu},
  year = {2025},
  month = apr,
  pages = {448--461},
  publisher = {Association for Computational Linguistics},
  address = {Albuquerque, New Mexico},
  doi = {10.18653/v1/2025.naacl-short.38},
  urldate = {2025-10-08},
  isbn = {979-8-89176-190-2}
}

@misc{das2025HallucinationsKey,
  title = {Hallucinations and {{Key Information Extraction}} in {{Medical Texts}}: {{A Comprehensive Assessment}} of {{Open-Source Large Language Models}}},
  shorttitle = {Hallucinations and {{Key Information Extraction}} in {{Medical Texts}}},
  author = {Das, Anindya Bijoy and Ahmed, Shibbir and Sakib, Shahnewaz Karim},
  year = {2025},
  month = aug,
  number = {arXiv:2504.19061},
  eprint = {2504.19061},
  primaryclass = {cs},
  publisher = {arXiv},
  doi = {10.48550/arXiv.2504.19061},
  urldate = {2025-10-07},
  archiveprefix = {arXiv}
}

@misc{fabbri2021SummEvalReevaluating,
  title = {{{SummEval}}: {{Re-evaluating Summarization Evaluation}}},
  shorttitle = {{{SummEval}}},
  author = {Fabbri, Alexander R. and Kry{\'s}ci{\'n}ski, Wojciech and McCann, Bryan and Xiong, Caiming and Socher, Richard and Radev, Dragomir},
  year = {2021},
  month = feb,
  number = {arXiv:2007.12626},
  eprint = {2007.12626},
  primaryclass = {cs},
  publisher = {arXiv},
  doi = {10.48550/arXiv.2007.12626},
  urldate = {2025-10-08},
  archiveprefix = {arXiv}
}

@inproceedings{fang2024UnderstandingFaithfulness,
  title = {Understanding {{Faithfulness}} and {{Reasoning}} of {{Large Language Models}} on {{Plain Biomedical Summaries}}},
  booktitle = {Findings of the {{Association}} for {{Computational Linguistics}}: {{EMNLP}} 2024},
  author = {Fang, Biaoyan and Dai, Xiang and Karimi, Sarvnaz},
  editor = {{Al-Onaizan}, Yaser and Bansal, Mohit and Chen, Yun-Nung},
  year = {2024},
  month = nov,
  pages = {9890--9911},
  publisher = {Association for Computational Linguistics},
  address = {Miami, Florida, USA},
  doi = {10.18653/v1/2024.findings-emnlp.578},
  urldate = {2025-10-07}
}

@misc{hegselmann2024DataCentricApproach,
  title = {A {{Data-Centric Approach To Generate Faithful}} and {{High Quality Patient Summaries}} with {{Large Language Models}}},
  author = {Hegselmann, Stefan and Shen, Shannon Zejiang and Gierse, Florian and Agrawal, Monica and Sontag, David and Jiang, Xiaoyi},
  year = {2024},
  month = jun,
  number = {arXiv:2402.15422},
  eprint = {2402.15422},
  primaryclass = {cs},
  publisher = {arXiv},
  doi = {10.48550/arXiv.2402.15422},
  urldate = {2025-03-07},
  archiveprefix = {arXiv}
}

@article{ji2023SurveyHallucination,
  title = {Survey of {{Hallucination}} in {{Natural Language Generation}}},
  author = {Ji, Ziwei and Lee, Nayeon and Frieske, Rita and Yu, Tiezheng and Su, Dan and Xu, Yan and Ishii, Etsuko and Bang, Yejin and Chen, Delong and Dai, Wenliang and Chan, Ho Shu and Madotto, Andrea and Fung, Pascale},
  year = {2023},
  month = dec,
  journal = {ACM Computing Surveys},
  volume = {55},
  number = {12},
  eprint = {2202.03629},
  primaryclass = {cs},
  pages = {1--38},
  issn = {0360-0300, 1557-7341},
  doi = {10.1145/3571730},
  urldate = {2025-10-08},
  archiveprefix = {arXiv}
}

@misc{kim2025MedicalHallucinations,
  title = {Medical {{Hallucinations}} in {{Foundation Models}} and {{Their Impact}} on {{Healthcare}}},
  author = {Kim, Yubin and Jeong, Hyewon and Chen, Shan and Li, Shuyue Stella and Lu, Mingyu and Alhamoud, Kumail and Mun, Jimin and Grau, Cristina and Jung, Minseok and Gameiro, Rodrigo and Fan, Lizhou and Park, Eugene and Lin, Tristan and Yoon, Joonsik and Yoon, Wonjin and Sap, Maarten and Tsvetkov, Yulia and Liang, Paul and Xu, Xuhai and Liu, Xin and McDuff, Daniel and Lee, Hyeonhoon and Park, Hae Won and Tulebaev, Samir and Breazeal, Cynthia},
  year = {2025},
  month = feb,
  number = {arXiv:2503.05777},
  eprint = {2503.05777},
  primaryclass = {cs},
  publisher = {arXiv},
  doi = {10.48550/arXiv.2503.05777},
  urldate = {2025-03-21},
  archiveprefix = {arXiv}
}

@inproceedings{maynez2020FaithfulnessFactuality,
  title = {On {{Faithfulness}} and {{Factuality}} in {{Abstractive Summarization}}},
  booktitle = {Proceedings of the 58th {{Annual Meeting}} of the {{Association}} for {{Computational Linguistics}}},
  author = {Maynez, Joshua and Narayan, Shashi and Bohnet, Bernd and McDonald, Ryan},
  editor = {Jurafsky, Dan and Chai, Joyce and Schluter, Natalie and Tetreault, Joel},
  year = {2020},
  month = jul,
  pages = {1906--1919},
  publisher = {Association for Computational Linguistics},
  address = {Online},
  doi = {10.18653/v1/2020.acl-main.173},
  urldate = {2025-10-08}
}

@article{tang2023EvaluatingLarge,
  title = {Evaluating Large Language Models on Medical Evidence Summarization},
  author = {Tang, Liyan and Sun, Zhaoyi and Idnay, Betina and Nestor, Jordan G. and Soroush, Ali and Elias, Pierre A. and Xu, Ziyang and Ding, Ying and Durrett, Greg and Rousseau, Justin F. and Weng, Chunhua and Peng, Yifan},
  year = {2023},
  month = aug,
  journal = {npj Digital Medicine},
  volume = {6},
  number = {1},
  pages = {158},
  publisher = {Nature Publishing Group},
  issn = {2398-6352},
  doi = {10.1038/s41746-023-00896-7},
  urldate = {2025-10-07},
  copyright = {2023 The Author(s)},
  langid = {english}
}

@article{veen2024AdaptedLarge,
  title = {Adapted {{Large Language Models Can Outperform Medical Experts}} in {{Clinical Text Summarization}}},
  author = {Veen, Dave Van and Uden, Cara Van and Blankemeier, Louis and Delbrouck, Jean-Benoit and Aali, Asad and Bluethgen, Christian and Pareek, Anuj and Polacin, Malgorzata and Reis, Eduardo Pontes and Seehofnerova, Anna and Rohatgi, Nidhi and Hosamani, Poonam and Collins, William and Ahuja, Neera and Langlotz, Curtis P. and Hom, Jason and Gatidis, Sergios and Pauly, John and Chaudhari, Akshay S.},
  year = {2024},
  month = apr,
  journal = {Nature Medicine},
  volume = {30},
  number = {4},
  eprint = {2309.07430},
  primaryclass = {cs},
  pages = {1134--1142},
  issn = {1078-8956, 1546-170X},
  doi = {10.1038/s41591-024-02855-5},
  urldate = {2025-03-09},
  archiveprefix = {arXiv}
}

@article{williams2025EvaluatingLarge,
  title = {Evaluating Large Language Models for Drafting Emergency Department Encounter Summaries},
  author = {Williams, Christopher Y. K. and Bains, Jaskaran and Tang, Tianyu and Patel, Kishan and Lucas, Alexa N. and Chen, Fiona and Miao, Brenda Y. and Butte, Atul J. and Kornblith, Aaron E.},
  year = {2025},
  month = jun,
  journal = {PLOS Digital Health},
  volume = {4},
  number = {6},
  pages = {e0000899},
  publisher = {Public Library of Science},
  issn = {2767-3170},
  doi = {10.1371/journal.pdig.0000899},
  urldate = {2025-10-07},
  langid = {english}
}

@article{zaretsky2024GenerativeArtificial,
  title = {Generative {{Artificial Intelligence}} to {{Transform Inpatient Discharge Summaries}} to {{Patient-Friendly Language}} and {{Format}}},
  author = {Zaretsky, Jonah and Kim, Jeong Min and Baskharoun, Samuel and Zhao, Yunan and Austrian, Jonathan and Aphinyanaphongs, Yindalon and Gupta, Ravi and Blecker, Saul B. and Feldman, Jonah},
  year = {2024},
  month = mar,
  journal = {JAMA network open},
  volume = {7},
  number = {3},
  pages = {e240357},
  issn = {2574-3805},
  doi = {10.1001/jamanetworkopen.2024.0357},
  langid = {english},
  pmcid = {PMC10928500},
  pmid = {38466307}
}

@misc{ji2023MitigatingHallucination,
  title = {Towards {{Mitigating Hallucination}} in {{Large Language Models}} via {{Self-Reflection}}},
  author = {Ji, Ziwei and Yu, Tiezheng and Xu, Yan and Lee, Nayeon and Ishii, Etsuko and Fung, Pascale},
  year = 2023,
  month = oct,
  number = {arXiv:2310.06271},
  eprint = {2310.06271},
  primaryclass = {cs},
  publisher = {arXiv},
  doi = {10.48550/arXiv.2310.06271},
  urldate = {2025-10-13},
  archiveprefix = {arXiv}
}

@inproceedings{koopman2023DrChatGPT,
  title = {Dr {{ChatGPT}} Tell Me What {{I}} Want to Hear: {{How}} Different Prompts Impact Health Answer Correctness},
  shorttitle = {Dr {{ChatGPT}} Tell Me What {{I}} Want to Hear},
  booktitle = {Proceedings of the 2023 {{Conference}} on {{Empirical Methods}} in {{Natural Language Processing}}},
  author = {Koopman, Bevan and Zuccon, Guido},
  editor = {Bouamor, Houda and Pino, Juan and Bali, Kalika},
  year = 2023,
  month = dec,
  pages = {15012--15022},
  publisher = {Association for Computational Linguistics},
  address = {Singapore},
  doi = {10.18653/v1/2023.emnlp-main.928},
  urldate = {2025-10-14}
}

@misc{lee2024RLAIFVs,
  title = {{{RLAIF}} vs. {{RLHF}}: {{Scaling Reinforcement Learning}} from {{Human Feedback}} with {{AI Feedback}}},
  shorttitle = {{{RLAIF}} vs. {{RLHF}}},
  author = {Lee, Harrison and Phatale, Samrat and Mansoor, Hassan and Mesnard, Thomas and Ferret, Johan and Lu, Kellie and Bishop, Colton and Hall, Ethan and Carbune, Victor and Rastogi, Abhinav and Prakash, Sushant},
  year = 2024,
  month = sep,
  number = {arXiv:2309.00267},
  eprint = {2309.00267},
  primaryclass = {cs},
  publisher = {arXiv},
  doi = {10.48550/arXiv.2309.00267},
  urldate = {2025-10-13},
  archiveprefix = {arXiv}
}

@misc{madaan2023SelfRefineIterative,
  title = {Self-{{Refine}}: {{Iterative Refinement}} with {{Self-Feedback}}},
  shorttitle = {Self-{{Refine}}},
  author = {Madaan, Aman and Tandon, Niket and Gupta, Prakhar and Hallinan, Skyler and Gao, Luyu and Wiegreffe, Sarah and Alon, Uri and Dziri, Nouha and Prabhumoye, Shrimai and Yang, Yiming and Gupta, Shashank and Majumder, Bodhisattwa Prasad and Hermann, Katherine and Welleck, Sean and Yazdanbakhsh, Amir and Clark, Peter},
  year = 2023,
  month = may,
  number = {arXiv:2303.17651},
  eprint = {2303.17651},
  primaryclass = {cs},
  publisher = {arXiv},
  doi = {10.48550/arXiv.2303.17651},
  urldate = {2025-10-13},
  archiveprefix = {arXiv}
}

@misc{mishra2024SYNFACEDITSynthetic,
  title = {{{SYNFAC-EDIT}}: {{Synthetic Imitation Edit Feedback}} for {{Factual Alignment}} in {{Clinical Summarization}}},
  shorttitle = {{{SYNFAC-EDIT}}},
  author = {Mishra, Prakamya and Yao, Zonghai and Vashisht, Parth and Ouyang, Feiyun and Wang, Beining and Mody, Vidhi Dhaval and Yu, Hong},
  year = 2024,
  month = oct,
  number = {arXiv:2402.13919},
  eprint = {2402.13919},
  primaryclass = {cs},
  publisher = {arXiv},
  doi = {10.48550/arXiv.2402.13919},
  urldate = {2025-10-13},
  archiveprefix = {arXiv}
}

@inproceedings{wan2024WhatEvidence,
  title = {What {{Evidence Do Language Models Find Convincing}}?},
  booktitle = {Proceedings of the 62nd {{Annual Meeting}} of the {{Association}} for {{Computational Linguistics}} ({{Volume}} 1: {{Long Papers}})},
  author = {Wan, Alexander and Wallace, Eric and Klein, Dan},
  editor = {Ku, Lun-Wei and Martins, Andre and Srikumar, Vivek},
  year = 2024,
  month = aug,
  pages = {7468--7484},
  publisher = {Association for Computational Linguistics},
  address = {Bangkok, Thailand},
  doi = {10.18653/v1/2024.acl-long.403},
  urldate = {2025-10-14}
}

@misc{llama3.1-8b-instruct,
  author       = {Meta},
  title        = {meta‑llama/Llama‑3.1‑8B‑Instruct},
  year         = {2024},
  howpublished = {Online; accessed 2025‑02‑21},
  url          = {https://huggingface.co/meta-llama/Llama-3.1-8B-Instruct}
}

@misc{llama3.2-3b-instruct,
  author       = {Meta},
  title        = {meta-llama/Llama-3.2-3B-Instruct},
  year         = {2024},
  howpublished = {Online; accessed 2026-01-05},
  url          = {https://huggingface.co/meta-llama/Llama-3.2-3B-Instruct}
}

@misc{gemma3-4b-it,
  author       = {Google},
  title        = {google/gemma-3-4b-it},
  year         = {2025},
  howpublished = {Online; accessed 2026-01-05},
  url          = {https://huggingface.co/google/gemma-3-4b-it}
}

@ARTICLE{Kraljevic2021-ln,
  title="Multi-domain clinical natural language processing with {MedCAT}: The Medical Concept Annotation Toolkit",
  author="Kraljevic, Zeljko and Searle, Thomas and Shek, Anthony and Roguski, Lukasz and Noor, Kawsar and Bean, Daniel and Mascio, Aurelie and Zhu, Leilei and Folarin, Amos A and Roberts, Angus and Bendayan, Rebecca and Richardson, Mark P and Stewart, Robert and Shah, Anoop D and Wong, Wai Keong and Ibrahim, Zina and Teo, James T and Dobson, Richard J B",
  journal="Artif. Intell. Med.",
  volume=117,
  pages="102083",
  month=jul,
  year=2021,
  issn="0933-3657",
  doi="10.1016/j.artmed.2021.102083"
}

@inproceedings{DBLP:conf/aaai/FlemingLHJRTBGS24,
  author       = {Scott L. Fleming and Alejandro Lozano and William J. Haberkorn and Jenelle A. Jindal and Eduardo Reis and Rahul Thapa and Louis Blankemeier and Julian Z. Genkins and Ethan Steinberg and Ashwin Nayak and Birju S. Patel and Chia{-}Chun Chiang and Alison Callahan and Zepeng Huo and Sergios Gatidis and Scott J. Adams and Oluseyi Fayanju and Shreya J. Shah and Thomas Savage and Ethan Goh and Akshay S. Chaudhari and Nima Aghaeepour and Christopher D. Sharp and Michael A. Pfeffer and Percy Liang and Jonathan H. Chen and Keith E. Morse and Emma P. Brunskill and Jason A. Fries and Nigam H. Shah},
  title        = {MedAlign: {A} Clinician-Generated Dataset for Instruction Following with Electronic Medical Records},
  booktitle    = {Thirty-Eighth {AAAI} Conference on Artificial Intelligence},
  year         = {2024},
  url          = {https://doi.org/10.1609/aaai.v38i20.30205},
  doi          = {10.1609/AAAI.V38I20.30205},
}

@misc{OpenAI2025GPT5,
  author       = {{OpenAI}},
  title        = {GPT-5 System Card},
  year         = {2025},
  howpublished = {\url{https://cdn.openai.com/gpt-5-system-card.pdf}},
  note         = {Official technical report describing the GPT-5 architecture and capabilities},
}

\newpage
\onecolumn
\appendix
\section{Additional Details}

\subsection{Summarization Prompt}
\begin{promptbox}[Summarization Prompt]
\begin{lstlisting}
<|system|>
You are a helpful assistant that helps patients understand their medical records.
<|end|>

<|user|>
You will be given some doctor's notes and you will need to summarize the patient's brief hospital course in one paragraph. Please only include key events and findings and avoid using medical jargons, and you MUST start the summary with "You were admitted".
{% if demonstrations %}
Here are some examples:

{% for example in demonstrations %}
DOCUMENT: 
{{ example.text }}

SUMMARY: 
{{ example.summary }}

{% endfor %}
{% endif %}
DOCUMENT: {{ context }}
<|end|>
\end{lstlisting}
\end{promptbox}

\subsection{Self-refinement Prompts}
\label{app:self-refinement}
\begin{promptbox}[Self-refinement revision without detectors Prompt]
\begin{lstlisting}
<|system|>
You are a clinical summarization assistant.
<|end|>

<|user|>
You will be given:
- A brief hospital course (SOURCE FACTS)
- A DRAFT SUMMARY that contains potential issues.  

Your task is to REVISE the DRAFT SUMMARY so that it is fully consistent with the SOURCE FACTS and suitable as an accurate after-visit summary.

Instructions:
1. Treat the SOURCE FACTS as authoritative.
2. Check for missing or incomplete information in the DRAFT SUMMARY compared to the SOURCE FACTS, especially:
   - Chief complaint / reason for visit  
   - Presenting symptoms  
   - Procedures performed  
   - Medications (new, changed, or discontinued)  
   - Vital signs  
   - Key laboratory or imaging findings
4. Do **not** invent or infer new diagnoses, medications, procedures, or dates.
5. Keep professional tone and structure suitable for a discharge or after-visit summary.
6. Prefer terms and values exactly as stated in the SOURCE FACTS.
7. If the DRAFT SUMMARY is missing clinically important info that IS present in the SOURCE FACTS (e.g. a key procedure or discharge medication), ADD it.
8. Always start the summary with "You were admitted" and refer to the patient as you / your.
9. Return only the revised summary text (no explanations or markup).

---
[SOURCE FACTS / BRIEF HOSPITAL COURSE]
{{ context }}

[DRAFT SUMMARY WITH FEEDBACK ANNOTATIONS]
{{ summary }}
<|end|>
\end{lstlisting}
\end{promptbox}

\begin{promptbox}[Self-refinement revision with detectors Prompt]
\begin{lstlisting}
<|system|>
You are a clinical summarization assistant.
<|end|>

<|user|>
You will be given:
- A brief hospital course (SOURCE FACTS)
- A DRAFT SUMMARY that contains feedback annotations marking potential issues.  
  Potentially incorrect or unsupported text is enclosed in <error> ... </error> tags.

Your task is to REVISE the DRAFT SUMMARY so that it is fully consistent with the SOURCE FACTS and suitable as an accurate after-visit summary.

Instructions:
1. Treat the SOURCE FACTS as authoritative.
2. For each segment wrapped in <error> ... </error>:
   - If the content is unsupported or contradicted by the SOURCE FACTS -> remove or correct it.
   - If the content is partially correct -> rewrite it using information from the SOURCE FACTS.
   - If the content is accurate -> keep it, but remove the <error> tags.
3. Check for missing or incomplete information in the DRAFT SUMMARY compared to the SOURCE FACTS, especially:
   - Chief complaint / reason for visit  
   - Presenting symptoms  
   - Procedures performed  
   - Medications (new, changed, or discontinued)  
   - Vital signs  
   - Key laboratory or imaging findings
4. Do **not** invent or infer new diagnoses, medications, procedures, or dates.
5. Keep professional tone and structure suitable for a discharge or after-visit summary.
6. Prefer terms and values exactly as stated in the SOURCE FACTS.
7. If the DRAFT SUMMARY is missing clinically important info that IS present in the SOURCE FACTS (e.g. a key procedure or discharge medication), ADD it.
8. Always start the summary with "You were admitted" and refer to the patient as you / your.
9. Return only the revised summary text (no explanations or markup).

---
[SOURCE FACTS / BRIEF HOSPITAL COURSE]
{{ context }}

[DRAFT SUMMARY WITH FEEDBACK ANNOTATIONS]
{{ summary_with_errors }}
<|end|>
\end{lstlisting}
\end{promptbox}

\subsection{Hallucination Detection}
\begin{promptbox}[Medalign zero-shot Prompt]
\begin{lstlisting}
<|system|>
You are a helpful assistant that helps patients understand their medical records.
<|end|>

<|user|>
You will be given a doctor's notes and a summary with potentially incorrectness. Your task is to identify spans with erroneous, contradictory, or unsupported facts in the summary, and label them using the <error> tag (e.g. <error>incorrect fact</error>). There could be more than one error in the summary. 
{% if demonstrations %}
Here are some examples:

{% for example in demonstrations %}
DOCUMENT: 
{{ example.context }}

ORIGINAL SUMMARY: 
{{ example.summary }}

SUMMARY WITH LABELED ERRORS:
{{ example.summary_with_errors }}

{% endfor %}
{% endif %}
Can you identify the errors for the following document and summary?
DOCUMENT: 
{{ context }}

ORIGINAL SUMMARY: 
{{ summary }}

SUMMARY WITH LABELED ERRORS:
<|end|>
\end{lstlisting}
\end{promptbox}

\begin{promptbox}[Medalign Chain-of-thought k-shot Prompt]
\begin{lstlisting}
<|system|>
You are a helpful assistant that helps patients understand their medical records.
<|end|>

<|user|>
We will present you with a pair of a brief hospital course (BHC) and a patient after visit summary (AVS). The AVS is also referred to as discharge summary. The BHC contains a detailed summary of the hospital stay written by medical service. It usually contains medical jargon, and it can follow different structures based on the hospital course and responsible medical specialty. The AVS summarizes the hospital stay for the patient in plain language. In practice, the BHC is not the only source of information to write the AVS. However, in our setting we treat the BHC as the only context for the summary.

For this labeling task, we are interested in errors in the AVS that are either unsupported by the BHC, contradict content in the BHC, or are wrong medical facts. We allow statements that contain general medical knowledge or advice that are often used in patient summaries. Most errors are due to unsupported facts, so we further distinguish those based on their specific content. This leads to the following error types or labels:
1. Unsupported facts, including condition/procedure/medication/time/location/number/name/word/other
2. Contradicted fact
3. Incorrect fact
And below is the detailed guideline, and we label error spans with the <error> tag (e.g. <error>incorrect fact</error>).

### Allowed General Medical Knowledge and Medical Advice
We allow general medical knowledge and advice that is often part of the AVS. Usually, these are information that are not specific for the hospital course given in the BHC. For example
- "Please take your medications as prescribed" contains no error even though the BHC does not contain this instruction because this is general medical advice.
- "If the symptoms get worse, please contact your doctor" contains no error even when the BHC does not contain this fact, since it is general medical knowledge that a doctor should be seen for worsening symptoms. 

### Determining Span of Errors
We label the smallest possible consecutive span that specifies the error given the BHC as a context. Removing further parts from the span would remove important information. A useful heuristic is to identify the minimal span that must be replaced to obtain a correct statement that is grammatically correct. For example
- "We performed an <error>esophageal-gastro-duodenoscopy (EGD).<error>" when no such procedure is reported in the BHC. The article "an" is not labeled as an error. When no procedure at all was performed "performed an esophageal-gastro-duodenoscopy (EGD)" should be labeled as error because there is no suitable substitute for "esophageal-gastro-duodenoscopy (EGD)".
- "After the surgery, we <error>transitioned you to oral oxycodone</error>." when the BHC contains no information for such a transition. If another medication transition is mentioned in the BHC and makes sense in this sentence only "oral oxycodone" should be labeled. If another oral medication transition is mentioned in the BHC only "oxycodone" should be labeled.
- "<error>Your symptoms responded well</error>." when no part of the sentence makes sense in the given context of the AVS.

### Dealing with Deidentified Information
The data contains deidentified information shown with "___" in the text. We always treat this as non-existent information. So, the annotators should not infer what the deidentified information could be. In general, deidentified fields in the AVS should not be labeled as errors. However, sometimes they belong to a wrong statement or clearly contain unsupported information (e.g., a doctor's name or phone numbers) that are not given in the BHC. In these cases, deidentified fields should be included in the error span. For example
- "Take ___ <error>200mg daily</error> and try to rest" when no such dosage information is provided in the BHC, but the statement to rest. The deidentified medication name is excluded from the error span.
- "Please avoid going up <error>more than ___ stairs</error> at a time" when restrictions for the number of stairs taken at a time are note mentioned in the BHC.
- "<error>Dr. ___ will follow up with you</error>" when no follow-up is mentioned in the BHC.
- "Please stop taking Aspirin <error>on ___</error>" when no stopping date is given in the BHC. 
- "Your RBC peaked <error>at ___ million</error>" if there is no hint of a specific red blood cell count given in the BHC.

### One Error per Span
To get reliable error counts a span should only contain a single error.
- "You received <error>Tylenol</error> and <error>Ciprofloxacin</error>" when there is no evidence in the BHC that the two medications were administered to the patient.
- "You have a <error>follow-up appointment with your PCP</error> and <error>your cardiologist</error>" when no such follow up is mentioned in the BHC. Both errors are labeled separately.

{% for example in demonstrations %}
### Example {{ loop.index }}

BHC:
{{ example.context }}

AVS:
{{ example.summary }}
{% if example.error_descriptions %}

ERRORS:
{{ example.error_descriptions }}

{% endif %}
AVS WITH ERRORS LABELED:
{{ example.summary_with_errors }}

{% endfor %}
{% if demonstrations %}

### Example {{ len(demonstrations) + 1 }}
{% endif %}

BHC:
{{ context }}

AVS:
{{ summary }}

ERROR:
<|end|>
\end{lstlisting}
\end{promptbox}

\begin{promptbox}[Medalign Chain-of-thought K-Shot with Class Explanation Prompt]
\begin{lstlisting}
<|system|>
You are a helpful assistant that helps patients understand their medical records.
<|end|>

<|user|>
We will present you with a pair of a brief hospital course (BHC) and a patient after visit summary (AVS). The AVS is also referred to as discharge summary. The BHC contains a detailed summary of the hospital stay written by medical service. It usually contains medical jargon, and it can follow different structures based on the hospital course and responsible medical specialty. The AVS summarizes the hospital stay for the patient in plain language. In practice, the BHC is not the only source of information to write the AVS. However, in our setting we treat the BHC as the only context for the summary.

## Instructions

For this labeling task, we are interested in errors in the AVS that are either unsupported by the BHC, contradict content in the BHC, or are wrong medical facts. We allow statements that contain general medical knowledge or advice that are often used in patient summaries. Most errors are due to unsupported facts, so we further distinguish those based on their specific content. This leads to the following error types or labels:
1. Unsupported facts, including condition/procedure/medication/time/location/number/name/word/other
2. Contradicted fact
3. Incorrect fact
And below is the detailed guideline, and we label error spans with the <error> tag (e.g. <error>incorrect fact</error>).

### Determining Span of Errors
We label the smallest possible consecutive span that specifies the error given the BHC as a context. Removing further parts from the span would remove important information. A useful heuristic is to identify the minimal span that must be replaced to obtain a correct statement that is grammatically correct. For example
- "We performed an <error>esophageal-gastro-duodenoscopy (EGD).<error>" when no such procedure is reported in the BHC. The article "an" is not labeled as an error. When no procedure at all was performed "performed an esophageal-gastro-duodenoscopy (EGD)" should be labeled as error because there is no suitable substitute for "esophageal-gastro-duodenoscopy (EGD)".
- "After the surgery, we <error>transitioned you to oral oxycodone</error>." when the BHC contains no information for such a transition. If another medication transition is mentioned in the BHC and makes sense in this sentence only "oral oxycodone" should be labeled. If another oral medication transition is mentioned in the BHC only "oxycodone" should be labeled.
- "<error>Your symptoms responded well</error>." when no part of the sentence makes sense in the given context of the AVS.

We allow general medical knowledge and advice that is often part of the AVS. Usually, these are information that are not specific for the hospital course given in the BHC. For example
- "Please take your medications as prescribed" contains no error even though the BHC does not contain this instruction because this is general medical advice.
- "If the symptoms get worse, please contact your doctor" contains no error even when the BHC does not contain this fact, since it is general medical knowledge that a doctor should be seen for worsening symptoms. 

We try to ignore grammatical errors in the BHC and AVS. If the original meaning can still be inferred (e.g. "medictaions" instead of "medications"), the most likely corrected form can be used. If the meaning cannot be inferred, they can be ignored in the BHC or labeled as Unsupported Other in the AVS.

If a sentence or phrase is repeated, then please treat it as you would any other sentence and highlight all errors (even if you did so in a previous sentence). For example
- "Please take Tylenol. Please take Tylenol" when Tylenol was prescribed in the BHC.
- "Limit your <error>use of stairs</error>. Please limit <error>use of stairs</error>" when movement was encouraged.

To get reliable error counts a span should only contain a single error.
- "You received <error>Tylenol</error> and <error>Ciprofloxacin</error>" when there is no evidence in the BHC that the two medications were administered to the patient.
- "You have a <error>follow-up appointment with your PCP</error> and <error>your cardiologist</error>" when no such follow up is mentioned in the BHC. Both errors are labeled separately.

### Dealing with Deidentified Information
The data contains deidentified information shown with "___" in the text. We always treat this as non-existent information. So, the annotators should not infer what the deidentified information could be. In general, deidentified fields in the AVS should not be labeled as errors. However, sometimes they belong to a wrong statement or clearly contain unsupported information (e.g., a doctor's name or phone numbers) that are not given in the BHC. In these cases, deidentified fields should be included in the error span. For example
- "Take ___ <error>200mg daily</error> and try to rest" when no such dosage information is provided in the BHC, but the statement to rest. The deidentified medication name is excluded from the error span.
- "Please avoid going up <error>more than ___ stairs</error> at a time" when restrictions for the number of stairs taken at a time are note mentioned in the BHC.
- "<error>Dr. ___ will follow up with you</error>" when no follow-up is mentioned in the BHC.
- "Please stop taking Aspirin <error>on ___</error>" when no stopping date is given in the BHC. 
- "Your RBC peaked <error>at ___ million</error>" if there is no hint of a specific red blood cell count given in the BHC.

### Error Types
In general, we ask for the most specific error that is applicable. If there is uncertainty which type applies, prefer the one mentioned first in the enumeration of all error types shown earlier. For instance, if the error contains an unsupported medication name, the Unsupported medication type should be used instead of the Unsupported name type. Here is a detailed description of the error types:
- `Unsupported Condition`: includes unsupported symptoms, diseases, or findings of the patient. For example
    - "You were found to have a <error>left clavicle fracture</error>" when no information was given for this condition in the BHC.
- `Unsupported Procedure`: includes any unsupported medical procedures. For example
    - "You had a <error>filter placed in your vein</error>" when no intervention with a filter was mentioned.
- `Unsupported Medication`: contains all errors related to unsupported medications. This includes medication classes, substances, routes, frequencies, and dosages. For example
    - "You were placed on <error>antibiotics</error>" when only blood thinners were prescribed.
- `Unsupported Time`: includes all errors for unsupported time or interval statements. For example
    - "Keep your arm in a sling for the <error>next 6 weeks</error>" when no specific duration is given.
- `Unsupported Location`: Locations include both unsupported physical places as well as regions of the patient. For example
    - "The patient was admitted to the <error>Acute Surgery Service</error>" when no admission location was provided in the BHC.
- `Unsupported Number`: any number either as digits or written that are unsupported. This also includes words such as "a" and "an". For example
    - "Your pacemaker rate was increased to <error>50</error>" when the rate of 50 is not given in the BHC.
- `Unsupported Name`: named entities that are not supported by the BHC. For example
    - "You were seen by the <error>interventional pulmonary service</error>" when no consult with this service was mentioned in the BHC.
- `Unsupported Word`: incorrect or inappropriate words or phrases which do not fit in any of the above types. For example
    - "We will send you home with a <error>drain</error> in place" when drain not mentioned in the BHC.
- `Unsupported Other`: If there is a mistake which clearly does not belong to any of the above categories, you may use this category as a last resort. We cannot give precise instructions because the "other" category is very broad.
- `Contradicted Fact`: This error type is independent of the content and contains all facts that clearly contradict information provided in the BHC. For example
    - "Your pacemaker rate was increased to <error>50</error>" when the context state a pacemaker rate of 40.
- `Incorrect Fact`: This error type is independent of the content and contains all facts that clearly contradict general medical knowledge or advice. For example
    - "We diagnosed a seizure, and you <error>can continue driving your car</error>" when no reason for allowing driving after a seizure is provided this contradict common medical knowledge.

{% for example in demonstrations %}
## Examples
### Example {{ loop.index }}

BHC:
{{ example.context }}

AVS:
{{ example.summary }}
{% if example.error_descriptions %}

ERRORS:
{{ example.error_descriptions }}

{% endif %}
AVS WITH ERRORS LABELED:
{{ example.summary_with_errors }}

{% endfor %}
{% if demonstrations %}

### Example {{ len(demonstrations) + 1 }}
{% endif %}
BHC:
{{ context }}

AVS:
{{ summary }}
<|end|>
\end{lstlisting}
\end{promptbox}

\begin{promptbox}[Medalign Chain-of-thought K-Shot with Class Explanation and Class Aware Prediction Prompt]
\begin{lstlisting}
<|system|>
You are a helpful assistant that helps patients understand their medical records.
<|end|>

<|user|>
We will present you with a pair of a brief hospital course (BHC) and a patient after visit summary (AVS). The AVS is also referred to as discharge summary. The BHC contains a detailed summary of the hospital stay written by medical service. It usually contains medical jargon, and it can follow different structures based on the hospital course and responsible medical specialty. The AVS summarizes the hospital stay for the patient in plain language. In practice, the BHC is not the only source of information to write the AVS. However, in our setting we treat the BHC as the only context for the summary.

## Instructions

For this labeling task, we are interested in errors in the AVS that are either unsupported by the BHC, contradict content in the BHC, or are wrong medical facts. We allow statements that contain general medical knowledge or advice that are often used in patient summaries. Most errors are due to unsupported facts, so we further distinguish those based on their specific content. This leads to the following error types or labels:
1. Unsupported facts, including condition/procedure/medication/time/location/number/name/word/other
2. Contradicted fact
3. Incorrect fact
And below is the detailed guideline, and we label error spans with the <error> tag (e.g. <error class="error_type">incorrect fact</error>).

### Determining Span of Errors
We label the smallest possible consecutive span that specifies the error given the BHC as a context. Removing further parts from the span would remove important information. A useful heuristic is to identify the minimal span that must be replaced to obtain a correct statement that is grammatically correct. For example
- "We performed an <error>esophageal-gastro-duodenoscopy (EGD).<error>" when no such procedure is reported in the BHC. The article "an" is not labeled as an error. When no procedure at all was performed "performed an esophageal-gastro-duodenoscopy (EGD)" should be labeled as error because there is no suitable substitute for "esophageal-gastro-duodenoscopy (EGD)".
- "After the surgery, we <error>transitioned you to oral oxycodone</error>." when the BHC contains no information for such a transition. If another medication transition is mentioned in the BHC and makes sense in this sentence only "oral oxycodone" should be labeled. If another oral medication transition is mentioned in the BHC only "oxycodone" should be labeled.
- "<error>Your symptoms responded well</error>." when no part of the sentence makes sense in the given context of the AVS.

We allow general medical knowledge and advice that is often part of the AVS. Usually, these are information that are not specific for the hospital course given in the BHC. For example
- "Please take your medications as prescribed" contains no error even though the BHC does not contain this instruction because this is general medical advice.
- "If the symptoms get worse, please contact your doctor" contains no error even when the BHC does not contain this fact, since it is general medical knowledge that a doctor should be seen for worsening symptoms. 

We try to ignore grammatical errors in the BHC and AVS. If the original meaning can still be inferred (e.g. "medictaions" instead of "medications"), the most likely corrected form can be used. If the meaning cannot be inferred, they can be ignored in the BHC or labeled as Unsupported Other in the AVS.

If a sentence or phrase is repeated, then please treat it as you would any other sentence and highlight all errors (even if you did so in a previous sentence). For example
- "Please take Tylenol. Please take Tylenol" when Tylenol was prescribed in the BHC.
- "Limit your <error>use of stairs</error>. Please limit <error>use of stairs</error>" when movement was encouraged.

To get reliable error counts a span should only contain a single error.
- "You received <error>Tylenol</error> and <error>Ciprofloxacin</error>" when there is no evidence in the BHC that the two medications were administered to the patient.
- "You have a <error>follow-up appointment with your PCP</error> and <error>your cardiologist</error>" when no such follow up is mentioned in the BHC. Both errors are labeled separately.

### Dealing with Deidentified Information
The data contains deidentified information shown with "___" in the text. We always treat this as non-existent information. So, the annotators should not infer what the deidentified information could be. In general, deidentified fields in the AVS should not be labeled as errors. However, sometimes they belong to a wrong statement or clearly contain unsupported information (e.g., a doctor's name or phone numbers) that are not given in the BHC. In these cases, deidentified fields should be included in the error span. For example
- "Take ___ <error>200mg daily</error> and try to rest" when no such dosage information is provided in the BHC, but the statement to rest. The deidentified medication name is excluded from the error span.
- "Please avoid going up <error>more than ___ stairs</error> at a time" when restrictions for the number of stairs taken at a time are note mentioned in the BHC.
- "<error>Dr. ___ will follow up with you</error>" when no follow-up is mentioned in the BHC.
- "Please stop taking Aspirin <error>on ___</error>" when no stopping date is given in the BHC. 
- "Your RBC peaked <error>at ___ million</error>" if there is no hint of a specific red blood cell count given in the BHC.

### Error Types
In general, we ask for the most specific error that is applicable. If there is uncertainty which type applies, prefer the one mentioned first in the enumeration of all error types shown earlier. For instance, if the error contains an unsupported medication name, the Unsupported medication type should be used instead of the Unsupported name type. Here is a detailed description of the error types:
- `Unsupported Condition`: includes unsupported symptoms, diseases, or findings of the patient. For example
    - "You were found to have a <error class="unsupported_condition">left clavicle fracture</error>" when no information was given for this condition in the BHC.
- `Unsupported Procedure`: includes any unsupported medical procedures. For example
    - "You had a <error class="unsupported_procedure">filter placed in your vein</error>" when no intervention with a filter was mentioned.
- `Unsupported Medication`: contains all errors related to unsupported medications. This includes medication classes, substances, routes, frequencies, and dosages. For example
    - "You were placed on <error class="unsupported_medication">antibiotics</error>" when only blood thinners were prescribed.
- `Unsupported Time`: includes all errors for unsupported time or interval statements. For example
    - "Keep your arm in a sling for the <error class="unsupported_time">next 6 weeks</error>" when no specific duration is given.
- `Unsupported Location`: Locations include both unsupported physical places as well as regions of the patient. For example
    - "The patient was admitted to the <error class="unsupported_location">Acute Surgery Service</error>" when no admission location was provided in the BHC.
- `Unsupported Number`: any number either as digits or written that are unsupported. This also includes words such as "a" and "an". For example
    - "Your pacemaker rate was increased to <error class="unsupported_number">50</error>" when the rate of 50 is not given in the BHC.
- `Unsupported Name`: named entities that are not supported by the BHC. For example
    - "You were seen by the <error class="unsupported_name">interventional pulmonary service</error>" when no consult with this service was mentioned in the BHC.
- `Unsupported Word`: incorrect or inappropriate words or phrases which do not fit in any of the above types. For example
    - "We will send you home with a <error class="unsupported_word">drain</error> in place" when drain not mentioned in the BHC.
- `Unsupported Other`: If there is a mistake which clearly does not belong to any of the above categories, you may use this category as a last resort. We cannot give precise instructions because the "other" category is very broad.
- `Contradicted Fact`: This error type is independent of the content and contains all facts that clearly contradict information provided in the BHC. For example
    - "Your pacemaker rate was increased to <error class="contradicted_fact">50</error>" when the context state a pacemaker rate of 40.
- `Incorrect Fact`: This error type is independent of the content and contains all facts that clearly contradict general medical knowledge or advice. For example
    - "We diagnosed a seizure, and you <error class="incorrect_fact">can continue driving your car</error>" when no reason for allowing driving after a seizure is provided this contradict common medical knowledge.

{% for example in demonstrations %}
## Examples
### Example {{ loop.index }}

BHC:
{{ example.context }}

AVS:
{{ example.summary }}
{% if example.error_descriptions %}

ERRORS:
{{ example.error_descriptions }}

{% endif %}
AVS WITH ERRORS LABELED:
{{ example.summary_with_errors }}

{% endfor %}
{% if demonstrations %}
### Example {{ len(demonstrations) + 1 }}

{% endif %}
BHC:
{{ context }}

AVS:
{{ summary }}
<|end|>
\end{lstlisting}
\end{promptbox}

\end{document}